\documentclass[11pt]{article}

\usepackage[preprint]{acl}

\usepackage{times}
\usepackage{latexsym}

\usepackage[T1]{fontenc}
\usepackage[utf8]{inputenc}

\usepackage{microtype}

\usepackage{inconsolata}

\usepackage{graphicx}

\usepackage{multirow}
\usepackage{mdframed}
\usepackage{amssymb}
\usepackage{booktabs}
\usepackage{float}
\usepackage{tikz}
\usetikzlibrary{arrows.meta}
\usepackage{amsmath}

\newenvironment{taskexample}{%
\vspace{.5em}
  \begin{mdframed}[
    backgroundcolor=gray!5,
    linewidth=0.4pt,
    roundcorner=2pt,
    innerleftmargin=6pt,
    innerrightmargin=6pt,
    innertopmargin=3pt,
    innerbottommargin=3pt,
    skipabove=4pt,
    skipbelow=4pt
  ]\footnotesize
}{%
  \end{mdframed}
  \vspace{.5em}
}

\title{LogicSkills: A Structured Benchmark for Formal Reasoning\\in Large Language Models}

\author{
Brian Rabern\textsuperscript{\normalfont 1,2} \and
Philipp Mondorf\textsuperscript{\normalfont 3,4} \and
Barbara Plank\textsuperscript{\normalfont 3,4}
\\
\textrm{\textsuperscript{1}}Niche, Inc., Bend, OR, USA \\
\textrm{\textsuperscript{2}}Oregon State University – Cascades, Bend, OR, USA \\
\textrm{\textsuperscript{3}}MaiNLP, LMU Munich, Germany \\
\textrm{\textsuperscript{4}}Munich Center for Machine Learning (MCML), Munich, Germany \\
{\tt brian.rabern@gmail.com} \hspace{2em}
{\tt p.mondorf@lmu.de} \hspace{2em}
{\tt b.plank@lmu.de}
}

\begin{document}
\maketitle

\begin{abstract}
Large language models perform well on many logical reasoning benchmarks, but it remains unclear which core logical skills they truly master. To address this, we introduce \textsc{LogicSkills}, a benchmark that isolates three fundamental logical skills: (i) \emph{formal symbolization}\textemdash{}translating premises into first-order logic; (ii) \emph{countermodel construction}\textemdash{}showing that an argument is logically invalid by constructing a finite countermodel; and (iii) \emph{validity assessment}\textemdash{}determining whether a conclusion follows from a set of premises. Items are drawn from the two-variable fragment of first-order logic without identity and are presented in both English and a Carrollian nonce-word language. All instances are solver-verified with Z3 for correctness and non-triviality. Across conventional instruction-tuned LLMs, performance is high on \emph{validity assessment} but substantially lower on \emph{formal symbolization} and \emph{countermodel construction}, highlighting that high task-level accuracy can mask weaknesses in core logical skills. In contrast, recent reasoning-tuned models perform strongly across all three tasks, suggesting a more systematic logical skill profile.
\end{abstract}

\section{Introduction}\label{sec:introduction}
Reasoning in large language models (LLMs) is often evaluated \emph{behaviorally}\textemdash{}by whether systems succeed on tasks that humans would solve through explicit reasoning\textemdash{}without presupposing any particular internal method~\citep{huang-chang-2023-towards, mondorf2024beyond, Chen2026}. In this setting, logical problems serve as external \emph{proxies} for deductive competence: success should reflect sensitivity to logical structure rather than reliance on superficial cues or world knowledge. However, common benchmarks of logical competence~\citep{yang2023logical} often conflate distinct logical subskills, making it difficult to determine which skills LLMs genuinely master. 

For instance, a model might correctly carry out a formal proof when the premises are already in symbolic form, yet fail to translate natural-language sentences into that form. Likewise, a model might apply modus ponens and other inference rules reliably yet falter when asked to construct a countermodel demonstrating that an argument is invalid. Proof-theoretic facility of this kind leaves model-theoretic
understanding untested. Existing benchmarks such as \textsc{ProofWriter}~\citep{tafjord-etal-2021-proofwriter} and \textsc{LogicBench}~\citep{parmar-etal-2024-logicbench} evaluate inference over templated natural language, bundling parsing and inference into a single composite score rather than isolating either.

\begin{figure}[!t]
 \includegraphics[width=1\linewidth]{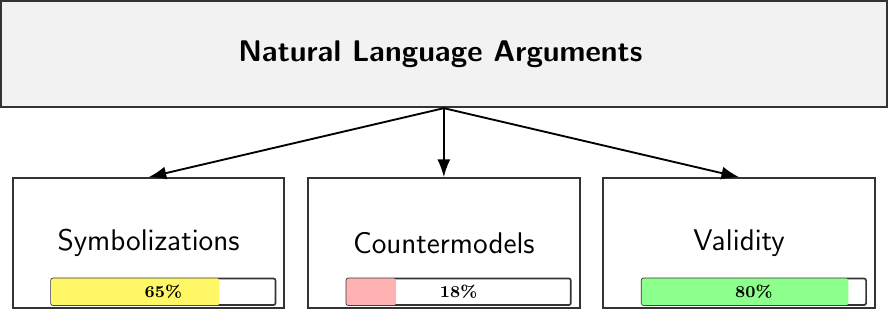}
\caption{Accuracy of standard instruction-tuned LLMs.}
\label{fig:LogicSkills-overview}
\end{figure}

To address this issue, we introduce \textsc{LogicSkills}, a benchmark for formal reasoning that disentangles fundamental logical skills and exposes failure modes that composite tasks may hide. Drawing on formal logic pedagogy, \textsc{LogicSkills} isolates three core logical skills: (i) \emph{formal symbolization} (mapping sentences to logical form), (ii) \emph{countermodel construction} (demonstrating invalidity via model-theoretic falsification), and (iii) \emph{validity assessment} (judging whether a conclusion follows from a set of premises). These skills are closely related: assessing validity typically requires mapping sentences to logical form and checking whether a countermodel exists in which the premises are true and the conclusion is false. \textsc{LogicSkills} is solver-verified (via Z3~\citep{10.1007/978-3-540-78800-3_24}) and includes a bilingual task design (English and Carrollian), with the Carrollian nonce-word condition isolating reasoning from prior semantic knowledge.

Systematic evaluations on \textsc{LogicSkills} across frontier open-weight and closed-weight LLMs reveal a notable pattern: conventional instruction-tuned LLMs perform strongly on \emph{validity assessment} but show substantial weaknesses in \emph{formal symbolization} and \emph{countermodel construction} (see Figure~\ref{fig:LogicSkills-overview}). This dissociation is notable, since a competent validity judgment must involve some representation of logical structure and some form of model-theoretic check, suggesting that such models may reach correct validity judgments without robustly symbolizing the premises or explicitly constructing countermodels. By contrast, reasoning-tuned LLMs perform strongly across all three skills, indicating a shift toward a more systematic logical skill profile. Together, our results call for more nuanced evaluations of logical reasoning in LLMs.

\section{Related Work}\label{sec:related_work}
Prior evaluations of logical reasoning in LLMs divide broadly into two traditions. 
\emph{Puzzle-based benchmarks} (e.g., knights-and-knaves, logic-grid, and minesweeper tasks) probe combinatorial reasoning often revealing brittleness under search or minor perturbations~\citep{mondorf-plank-2024-liar,lin2025zebralogic,giadikiaroglou-etal-2024-puzzle,Li_2024}. 
\emph{Rule- or proof-oriented suites} such as \textsc{ProofWriter}, \textsc{LogicBench}, and \textsc{LogicAsker} assess inference templates across propositional and first-order logic~\citep{tafjord-etal-2021-proofwriter,parmar-etal-2024-logicbench,han-etal-2024-folio,wan-etal-2024-logicasker}. 

\textsc{LogicSkills} complements both traditions by offering a structured, skill-level evaluation framework rather than a single composite reasoning task.

\section{Benchmark Overview}\label{sec:benchmark}
We build a compositional generator that produces grammatical natural-language sentences paired with formulas in the two-variable fragment of first-order logic (FO$^2$, without identity). We choose FO$^2$ since this logic retains decidability and supports finite satisfiability while being expressive enough to capture nontrivial reasoning~\citep{scott1962decision,mortimer1975}. Every formula is realized twice: (i) in controlled English and (ii) in a Carroll-style nonsense lexicon~\citep{carroll1871lookingglass}. The latter provides a structurally identical surface form that removes real-world associations and isolates pure logical competence. For example, the sentences ``A human chased a donkey that chased it'' and ``A borogove snicker-snacked a tove that snicker-snacked it'' both map to the formula:
\begin{equation}\label{eq:fol}
\exists x \exists y \big((Nx \wedge My) \wedge (Rxy \wedge Ryx)\big)
\end{equation}

\noindent By varying quantifier structure, predicate choice, name assignments, connectives, and recursive clausal patterns, the generator can produce a wide range of sentences in both English and Carrollian.

Because every sentence is paired with an FO$^2$ formula, we leverage SMT solvers, whose search procedures descend from
\citet{DavisPutnam1960}\textemdash{}in particular the state-of-the-art theorem prover Z3~\citep{10.1007/978-3-540-78800-3_24}\textemdash{}to compute core semantic properties: (i) whether two formulas are equivalent, (ii) whether a set of formulas is jointly satisfiable, (iii) whether one formula follows from others, and (iv) whether a proposed finite model satisfies (or falsifies) a given argument. This enables large-scale, correctness-guaranteed construction and verification of all deductive instances in the benchmark.

To construct arguments, we sample sets of 3--5 jointly satisfiable premises and use Z3 to search for semantically relevant, non-trivial conclusions that logically follow (valid). For each valid conclusion, we additionally generate 5 structurally similar distractor conclusions that do not follow (invalid). Appendix~\ref{app:pipeline} provides further details about the dataset and problem construction pipeline.

\subsection{Task types}\label{subsec:task_types}
We instantiate three task types, each aligned with a canonical logical skill \cite{kalishMontague1964}. Exact prompts can be found in Appendix \ref{sec:appendix_prompts}.

\paragraph{Formal symbolization} Given a sentence in English/Carroll and a fixed symbol key, models must output a single well-formed formula capturing the sentence using only the provided predicates, constants, and standard logical operators (i.e., $\lnot,\ \land,\ \lor,\ \rightarrow,\ \leftrightarrow,\ \forall,\ \exists$
).

\begin{taskexample}
\textbf{Input:}\\[2pt]
\hspace*{1em}Sentence: ``All raths are uffish.''\\
\hspace*{1em}Key: $F$: ``$x$ is a rath'',\quad $G$: ``$x$ is uffish''.\\[2pt]

\hrule\vspace{4pt}

\noindent \textbf{Output:} $\forall x\,(Fx \rightarrow Gx)$
\end{taskexample}

\paragraph{Countermodel construction} For an invalid argument, models must provide a finite structure (on a fixed small domain) that makes all premises true and the conclusion false.\footnote{Every item is verified to admit a countermodel for the given domain; see Appendix~\ref{app:pipeline}.}

\begin{taskexample}
\textbf{Input:}\\[2pt]
\hspace*{1em}Invalid Argument: $\forall x\,(Fx\rightarrow Gx),\; Ga \vDash Fa$.\\
\hspace*{1em}Domain: $\{0,1\}$\\[2pt]

\hrule\vspace{4pt}

\noindent \textbf{Output:} $F=\{0\}, G=\{0,1\}, a=1$.
\end{taskexample}

\paragraph{Validity assessment} Given premises and six candidate conclusions in English/Carroll, models must decide which conclusion(s), if any, \emph{must} follow.

\begin{taskexample}
\textbf{Input:}\\[2pt]
\hspace*{1em}Premises: ``All raths are uffish'';\; ``Zindle is a rath.''\\
\hspace*{1em}Candidate Conclusions:\\
\hspace*{1.5em}1.\; Zindle isn't uffish.\\
\hspace*{1.5em}2.\; Only raths are uffish.\\
\hspace*{1.5em}3.\; Zindle is uffish.\\
\hspace*{1.5em}4.\; No raths are uffish.\\
\hspace*{1.5em}5.\; Zindle snicker-snacked a rath.\\
\hspace*{1.5em}6.\; Some rath is not uffish.\\[2pt]

\hrule\vspace{4pt}

\noindent \textbf{Output:} 3
\end{taskexample}

\section{Evaluation Setup}\label{sec:evaluation}
The compositional data generator described in Section~\ref{sec:benchmark} is fully programmatic and supports the creation of a large number of unique problem instances for each task type. For example, we generate over 200k problem instances for the PEFT experiments in Section~\ref{sec:peft}. For the main evaluation, however, we use a fixed set of $1{,}500$ problem instances to keep evaluation costs manageable while ensuring broad coverage across tasks. Specifically, our evaluation set comprises $600$ problems from \emph{formal symbolization}, $600$ from \emph{validity assessment}, and $300$ from \emph{countermodel construction}. Symbolization and validity are evaluated bilingually ($300$ problems in English and $300$ in Carroll), whereas countermodel construction is language-neutral. Code for dataset generation, task construction, and model evaluation is publicly available at \href{https://github.com/brianrabern/LogicSkills}{https://github.com/brianrabern/LogicSkills}.

We evaluate a range of open- and closed-weight LLMs: \emph{Llama-3.1-8B}, \emph{Llama-3.1-70B}, \emph{Qwen2.5-Math-72B}, \emph{Claude-3.7-Sonnet}, \emph{GPT-4o}, \emph{Gemini-2.5-Flash}, \emph{DeepSeek-Chat}, and \emph{Phi-4}. We further assess two reasoning-tuned models: \emph{Qwen3-32B} and \emph{OpenAI-o3}. All models are evaluated with greedy decoding. The validity and countermodel prompts permit any solution strategy and place no constraint on intermediate reasoning; only the symbolization prompt restricts the response to a single formula, to ensure parseability (Appendix~\ref{sec:appendix_prompts}). Responses are processed in two stages. First, an extractor LLM (GPT-4o) cleans and normalizes raw completions.\footnote{A meta-evaluation shows 98–99\% extraction accuracy; see Appendix~\ref{app:meta-eval} for more details.} Second, task-specific verification procedures check correctness.

\paragraph{Symbolization} Normalized formulas are first checked for an exact match; if necessary, they are parsed and repaired. Cosmetic variants (e.g., $\wedge$ as \texttt{\&}, $\neg$ as $\sim$, $\rightarrow$ as $\supset$, or $R(x,y)$ vs.\ $Rxy$) are handled via  the extractor LLM. Final parses are validated for logical equivalence using Z3.

\paragraph{Countermodels} Candidate models are normalized into domain, constants, and predicate extensions. The system enforces \emph{symbol completeness} and validates type/domain consistency (integer domains, correct arities). Candidate models are then translated into SMT-LIB, merged with the negated argument, and checked for satisfiability with Z3.

\paragraph{Validity} Extracted index sets are compared against the gold sets under set equality (see Appendix~\ref{app:eval}).

Notably, the evaluation for symbolization and countermodels is semantic (via Z3) rather than string-based: symbolizations are judged by logical equivalence, and countermodels by whether they satisfy the premises and falsify the conclusion.

\begin{figure*}[t]
  \centering
  \includegraphics[width=0.8\linewidth]{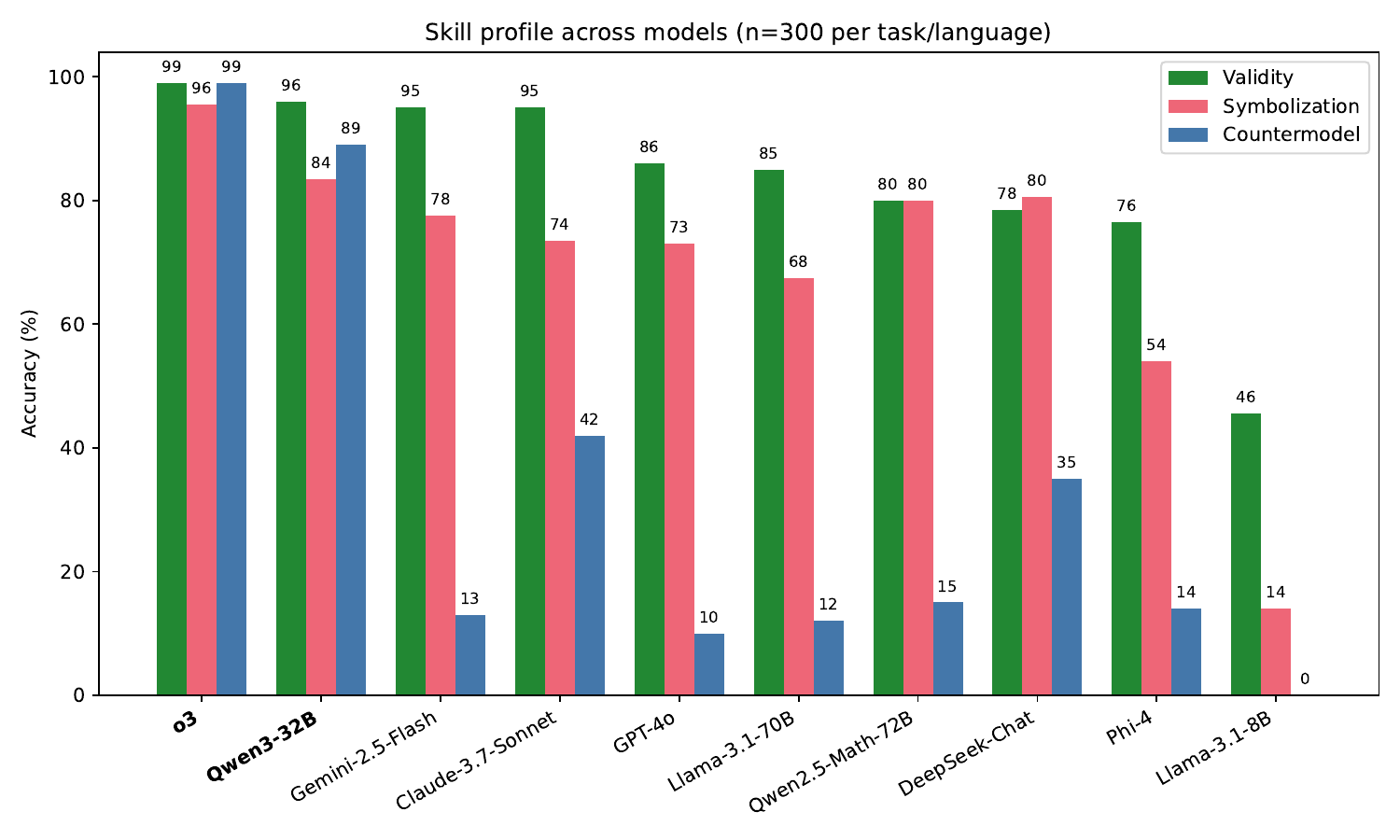}
  \caption{\textbf{Skill profile across models} (accuracy \%; $n{=}300$ per task/language).
  Grouped bars show \emph{Validity}, \emph{Symbolization}, and \emph{Countermodel}. Most models score highly on validity but lag on subskills, especially countermodels.}
  \label{fig:skill-profile}
\end{figure*}

\section{Results}\label{sec:results}
Figure~\ref{fig:skill-profile} illustrates model accuracy across the three task types described in Section~\ref{sec:benchmark}. A clear pattern emerges among instruction-tuned models: performance on \emph{validity assessment} is high, whereas performance on \emph{formal symbolization} and especially \emph{countermodel construction} remains substantially weaker. For example, GPT-4o achieves $86.3\%$ on \emph{validity assessment} but
only $10.0\%$ on \emph{countermodel construction}; Gemini-2.5-Flash reaches $95.2\%$ versus $13.3\%$, respectively. While accuracy on \emph{validity assessment} clusters near ceiling, suggesting that premise--conclusion classification is largely mastered under the given conditions, \emph{formal symbolization} remains a consistent bottleneck, reflecting persistent difficulty in mapping natural-language statements to formal representations. The \emph{countermodel construction} task is the most discriminative, revealing substantial gaps in model-theoretic reasoning, with most conventional LLMs failing to generate consistent falsifying structures. Taken together, these results suggest deficits in fundamental logical subskills that prior high-level composite tasks may have obscured.

In contrast, reasoning-tuned models such as \emph{Qwen3-32B} and \emph{o3} exhibit strong performance across all three task types. For instance, \emph{o3} achieves $98.8\%$ on \emph{validity assessment}, $98.7\%$ on \emph{countermodel construction}, and $95.3\%$ on \emph{formal symbolization}. This pattern suggests that reasoning-tuned models may be
moving toward more comprehensive logical competence.

\subsection{Bilingual evaluation}
Perhaps surprisingly, bilingual evaluation shows only minor differences: averaging across models, English and Carroll results are closely aligned (validity $83.0\%$ vs.\ $84.4\%$; symbolization $70.7\%$ vs.\ $69.2\%$), with a maximum difference of $6.7$ percentage points across all
conditions. This null result suggests that the models are not relying primarily on semantic heuristics when processing the syntactic structure of the arguments. For an overview of accuracy values per model, task, and language, we refer to Table~\ref{tab:full_overview} in the Appendix.

\subsection{Error analysis}
Error analyses across tasks reveal distinct but converging weaknesses in model reasoning.  
In the \emph{formal symbolization} task, over half of the 1,803 incorrect outputs (50.2\%) are semantically incorrect despite syntactic well-formedness, while the remainder show structural issues such as predicate-arity mismatches or ill-formed parentheses. In the \emph{countermodel construction} task, most erroneous outputs (72.9\% of 2,020) are structurally well-formed but fail to falsify the argument—typically because one or more premises are false—while the rest (27.1\%) reflect missing interpretations or malformed predicate definitions.  
In the \emph{validity assessment} task, a majority of the 980 incorrect outputs (58.4\%) include the correct answer alongside additional invalid options, suggesting difficulty in eliminating plausible distractors. Further details about the error analysis procedure can be found in Appendix~\ref{app:err_analysis}.

\subsection{Subskill Transfer via PEFT}\label{sec:peft}
In principle, skill at \emph{formal symbolization} and \emph{countermodel construction} should reinforce \emph{validity assessment}: to judge validity, a model must map sentences to their logical form and evaluate whether any counterexample structure can satisfy the premises while falsifying the conclusion. To assess potential cross-skill transfer, we conduct a fine-tuning study. We fine-tune \emph{Llama-3.2-3B-Instruct} with LoRA adapters on $100$k \emph{symbolization} and $100$k \emph{countermodel} examples (targets generated via tpg; \citealp{schwarz2025tpg}), using a single epoch with AdamW and standard regularization; evaluation items were held out of the training data, and the fine-tuned models are evaluated on the English portion of each task ($n{=}300$).

We observe large \emph{in-task} gains---symbolization rises from $8.3\%$ to $95.0\%$ and countermodel construction from $0.0\%$ to $23.0\%$---but no transfer to \emph{validity assessment} (Table~\ref{tab:transfer-matrix}). Validity accuracy declines under joint training ($-15.3\%$), but this is not a decline in judgment \emph{per se}: fine-tuning sharply reduces the proportion of responses that resolve into a selection among the candidate conclusions ($97.3\%$ for the base model, $61.0\%$ under joint training),\footnote{For example, rather than selecting among options 1-6, jointly fine-tuned responses often either reprint the candidate conclusions verbatim with little commentary, or devolve into repetitive step-numbered chains (since $p$, $q$ (1); since $q$, $r$ (2); since $q$, $s$ (3) ...) that recycle the same templates without ever selecting one of the six candidate conclusions; in both cases our extractor returns \emph{Indeterminate}.} while accuracy among responses that do resolve is essentially unchanged ($37.7\%$ base vs.\ $35.0\%$ joint). Fine-tuned models carry the response format of the training tasks into the validity prompt, producing structured derivations that never close on an answer. The subskills are thus acquired in isolation, without composing into a workflow the model can deploy on the validity task. The pattern parallels a student trained on translation and model construction but never shown how those skills bear on assessing an argument: the components are present, but nothing connects them to the task.

\begin{table}[h]
\centering
\small
\setlength{\tabcolsep}{4pt}
\renewcommand{\arraystretch}{0.95}
\begin{tabular}{lrrrrrr}
\toprule
& \multicolumn{2}{c}{\textbf{Sym.}} & \multicolumn{2}{c}{\textbf{Counter}} & \multicolumn{2}{c}{\textbf{Val.}}\\
\cmidrule(lr){2-3}\cmidrule(lr){4-5}\cmidrule(lr){6-7}
\textbf{Train Data} & Acc. & $\Delta$ & Acc. & $\Delta$ & Acc. & $\Delta$\\
\midrule
Base            & \phantom{0}8.3 & ---   & \phantom{0}0.0 & ---   & 36.7 & ---\\
Symb.\ (100k)   & 95.0 & +86.7 & \phantom{0}0.0 & \phantom{0}+0.0 & 39.7 & \phantom{0}+3.0\\
Counter (100k)  & \phantom{0}4.3 & \phantom{0}--4.0 & 23.0 & +23.0 & 32.0 & \phantom{0}--4.7\\
Joint (200k)    & 95.7 & +87.3 & 22.7 & +22.7 & 21.3 & --15.3\\
\bottomrule
\end{tabular}
\caption{\label{tab:transfer-matrix}\emph{Llama-3.2-3B-Instruct} accuracy (Acc. \%) and change ($\Delta$ \%) of the base model vs.\ after fine-tuning ($n{=}300$ per cell).}
\end{table}

\section{Conclusion}
From Leibniz’s (\citeyear{Leibniz1685}) call to ``let us calculate'' to Frege's \emph{Begriffsschrift} (\citeyear{Frege1879}), the project of logic has rested on the conviction that reasoning becomes reliable only when it is translated into a formal system, where validity is established through explicit rules. The evaluations on \textsc{LogicSkills} presented in this work suggest that conventional LLMs can achieve high accuracy on logical tasks without the explicit symbolic and model-based procedures that underpin standard logical methods. At the same time, recent reasoning-tuned models exhibit a broader and more systematic logical skill profile, pointing to a promising direction for future work.

\section*{Limitations}

While our study introduces a novel benchmark for formal reasoning that disentangles core logical skills and exposes failure modes that composite tasks may hide, there are several limitations that future work can address. Although the evaluation benchmark reported here is fixed in size, the underlying generation pipeline is fully programmatic and supports large-scale expansion; for example, we generate an additional 200k training instances for the PEFT experiments (\S \ref{sec:peft}) using the same infrastructure.

\paragraph{Logical and linguistic scope} \textsc{LogicSkills} evaluates reasoning within a deliberately restricted logical and linguistic scope. All tasks are drawn from the two-variable fragment of first-order logic without identity, excluding richer logical resources such as full first-order logic, identity, or modal operators. Consequently, results may not generalize to reasoning in more expressive logical systems. The benchmark also relies on a controlled fragment of English and a synthetically generated Carrollian language. While this design isolates logical structure, it limits linguistic diversity and does not capture many sources of difficulty present in naturally occurring language. 

\paragraph{Evaluation procedure} Model performance is additionally sensitive to prompting and evaluation protocols. In the symbolization and countermodel tasks, responses must contain logical content that can be interpreted as formulas or structures for semantic verification. However, evaluation does not depend on a fixed output template: responses are normalized, parsed, and repaired prior to verification, and cosmetic or notational variants are tolerated. Correctness is ultimately determined by logical equivalence checks and model-theoretic verification. Nevertheless, performance may still partly reflect a model's ability to produce interpretable logical content rather than reasoning ability alone. This also applies to the PEFT results: the fine-tuning data contained only symbolization and countermodel examples, with no validity or general instruction data mixed in, so drift toward the training tasks' response format is expected.

\paragraph{Behavior vs. Mechanism}
\textsc{LogicSkills} evaluates reasoning behaviorally. While it reveals systematic differences across logical subskills, it does not directly address the internal representations or mechanisms by which models arrive at their answers \citep{chalmers2025interpretability,stolfo2023mechanistic}.  Whether reasoning-oriented models that pass all three tasks are genuinely approximating model-theoretic reasoning or achieving this through more powerful but equally opaque means remains an open question for mechanistic investigation.

Within the scope of this paper, these limitations are largely intentional design choices, reflecting a tradeoff between expressivity and reliable, solver-verifiable evaluation.

\section*{Acknowledgments}
We would like to thank the anonymous reviewers for their comments and suggestions. Furthermore, we acknowledge the support for BP through the ERC Consolidator Grant DIALECT 101043235.

\bibliography{custom}

@inproceedings{wan-etal-2024-logicasker,
  title = {LogicAsker: Evaluating and Improving the Logical Reasoning Ability of Large Language Models},
  author = {Wan, Yuxuan and Wang, Wenxuan and Yang, Yiliu and Yuan, Youliang and Huang, Jen-tse and He, Pinjia and Jiao, Wenxiang and Lyu, Michael},
  booktitle = {Proceedings of the 2024 Conference on Empirical Methods in Natural Language Processing},
  pages = {2124--2155},
  address = {Miami, Florida, USA},
  publisher = {Association for Computational Linguistics},
  year = {2024},
  doi = {10.18653/v1/2024.emnlp-main.128}
}

@inproceedings{giadikiaroglou-etal-2024-puzzle,
    title = "Puzzle Solving using Reasoning of Large Language Models: A Survey",
    author = "Giadikiaroglou, Panagiotis  and
      Lymperaiou, Maria  and
      Filandrianos, Giorgos  and
      Stamou, Giorgos",
    editor = "Al-Onaizan, Yaser  and
      Bansal, Mohit  and
      Chen, Yun-Nung",
    booktitle = "Proceedings of the 2024 Conference on Empirical Methods in Natural Language Processing",
    month = nov,
    year = "2024",
    address = "Miami, Florida, USA",
    publisher = "Association for Computational Linguistics",
    url = "https://aclanthology.org/2024.emnlp-main.646/",
    doi = "10.18653/v1/2024.emnlp-main.646",
    pages = "11574--11591"
}

@inproceedings{stolfo2023mechanistic,
  title={A Mechanistic Interpretation of Arithmetic Reasoning in Language Models using Causal Mediation Analysis},
  author={Stolfo, Alessandro and Belinkov, Yonatan and Sachan, Mrinmaya},
  booktitle={Proceedings of the 2023 Conference on Empirical Methods in Natural Language Processing},
  year={2023}
}

@book{Frege1879,
  author    = {Gottlob Frege},
  title     = {Begriffsschrift: eine der arithmetischen nachgebildete Formelsprache des reinen Denkens},
  year      = {1879},
  publisher = {Verlag von Louis Nebert},
  address   = {Halle},
  note      = {English translation in Jean van Heijenoort (ed.), \textit{From Frege to G{\"o}del: A Source Book in Mathematical Logic, 1879--1931}, Harvard University Press, 1967, pp. 1--82.}
}

@incollection{Leibniz1685,
  author    = {Gottfried Wilhelm Leibniz},
  title     = {The Art of Discovery},
  booktitle = {Leibniz: Selections},
  editor    = {Philip P. Wiener},
  publisher = {Charles Scribner's Sons},
  address   = {New York},
  year      = {1685},
  note      = {Translation by P.~P. Wiener, 1951, p.~51; Latin in
               \emph{Opuscules et Fragments In\'edits de Leibniz},
               ed.\ L.~Couturat, Paris: F\'elix Alcan, 1903, p.~176.}
}

@inproceedings{han-etal-2024-folio,
    title = "{FOLIO}: Natural Language Reasoning with First-Order Logic",
    author = "Han, Simeng  and
      Schoelkopf, Hailey  and
      Zhao, Yilun  and
      Qi, Zhenting  and
      Riddell, Martin  and
      Zhou, Wenfei  and
      Coady, James  and
      Peng, David  and
      Qiao, Yujie  and
      Benson, Luke  and
      Sun, Lucy  and
      Wardle-Solano, Alexander  and
      Szab{\'o}, Hannah  and
      Zubova, Ekaterina  and
      Burtell, Matthew  and
      Fan, Jonathan  and
      Liu, Yixin  and
      Wong, Brian  and
      Sailor, Malcolm  and
      Ni, Ansong  and
      Nan, Linyong  and
      Kasai, Jungo  and
      Yu, Tao  and
      Zhang, Rui  and
      Fabbri, Alexander  and
      Kryscinski, Wojciech Maciej  and
      Yavuz, Semih  and
      Liu, Ye  and
      Lin, Xi Victoria  and
      Joty, Shafiq  and
      Zhou, Yingbo  and
      Xiong, Caiming  and
      Ying, Rex  and
      Cohan, Arman  and
      Radev, Dragomir",
    editor = "Al-Onaizan, Yaser  and
      Bansal, Mohit  and
      Chen, Yun-Nung",
    booktitle = "Proceedings of the 2024 Conference on Empirical Methods in Natural Language Processing",
    month = nov,
    year = "2024",
    address = "Miami, Florida, USA",
    publisher = "Association for Computational Linguistics",
    url = "https://aclanthology.org/2024.emnlp-main.1229/",
    doi = "10.18653/v1/2024.emnlp-main.1229",
    pages = "22017--22031"
}

@book{carroll1871lookingglass,
  author    = {Lewis Carroll},
  title     = {Through the Looking-Glass, and What Alice Found There},
  year      = {1871},
  publisher = {Macmillan and Co.},
  address   = {London},
  note      = {Chapter 1, pp. 21--24}
}

@misc{chalmers2025interpretability,
      title={Propositional Interpretability in Artificial Intelligence}, 
      author={David J. Chalmers},
      year={2025},
      eprint={2501.15740},
      archivePrefix={arXiv},
      primaryClass={cs.AI},
      url={https://arxiv.org/abs/2501.15740}, 
}

@misc{schwarz2025tpg,
  author       = {Wolfgang Schwarz},
  title        = {Tree Proof Generator (TPG)},
  year         = {2025},
  howpublished = {\url{https://github.com/wo/tpg}},
  note         = {See also \url{https://www.umsu.de/trees/}}
}

@inproceedings{mondorf-plank-2024-liar,
    title = "Liar, Liar, Logical Mire: A Benchmark for Suppositional Reasoning in Large Language Models",
    author = "Mondorf, Philipp  and
      Plank, Barbara",
    editor = "Al-Onaizan, Yaser  and
      Bansal, Mohit  and
      Chen, Yun-Nung",
    booktitle = "Proceedings of the 2024 Conference on Empirical Methods in Natural Language Processing",
    month = nov,
    year = "2024",
    address = "Miami, Florida, USA",
    publisher = "Association for Computational Linguistics",
    url = "https://aclanthology.org/2024.emnlp-main.404/",
    doi = "10.18653/v1/2024.emnlp-main.404",
    pages = "7114--7137"
}

@inproceedings{Li_2024,
   title={Assessing Logical Puzzle Solving in Large Language Models: Insights from a Minesweeper Case Study},
   url={http://dx.doi.org/10.18653/v1/2024.naacl-long.4},
   DOI={10.18653/v1/2024.naacl-long.4},
   booktitle={Proceedings of the 2024 Conference of the North American Chapter of the Association for Computational Linguistics: Human Language Technologies (Volume 1: Long Papers)},
   publisher={Association for Computational Linguistics},
   author={Li, Yinghao and Wang, Haorui and Zhang, Chao},
   year={2024},
   pages={59–81} }

@inproceedings{tafjord-etal-2021-proofwriter,
    title = "{P}roof{W}riter: Generating Implications, Proofs, and Abductive Statements over Natural Language",
    author = "Tafjord, Oyvind  and
      Dalvi, Bhavana  and
      Clark, Peter",
    editor = "Zong, Chengqing  and
      Xia, Fei  and
      Li, Wenjie  and
      Navigli, Roberto",
    booktitle = "Findings of the Association for Computational Linguistics: ACL-IJCNLP 2021",
    month = aug,
    year = "2021",
    address = "Online",
    publisher = "Association for Computational Linguistics",
    url = "https://aclanthology.org/2021.findings-acl.317/",
    doi = "10.18653/v1/2021.findings-acl.317",
    pages = "3621--3634"
}

@inproceedings{
lin2025zebralogic,
title={ZebraLogic: On the Scaling Limits of {LLM}s for Logical Reasoning},
author={Bill Yuchen Lin and Ronan Le Bras and Kyle Richardson and Ashish Sabharwal and Radha Poovendran and Peter Clark and Yejin Choi},
booktitle={Forty-second International Conference on Machine Learning},
year={2025},
url={https://openreview.net/forum?id=sTAJ9QyA6l}
}

@inproceedings{parmar-etal-2024-logicbench,
    title = "{L}ogic{B}ench: Towards Systematic Evaluation of Logical Reasoning Ability of Large Language Models",
    author = "Parmar, Mihir  and
      Patel, Nisarg  and
      Varshney, Neeraj  and
      Nakamura, Mutsumi  and
      Luo, Man  and
      Mashetty, Santosh  and
      Mitra, Arindam  and
      Baral, Chitta",
    editor = "Ku, Lun-Wei  and
      Martins, Andre  and
      Srikumar, Vivek",
    booktitle = "Proceedings of the 62nd Annual Meeting of the Association for Computational Linguistics (Volume 1: Long Papers)",
    month = aug,
    year = "2024",
    address = "Bangkok, Thailand",
    publisher = "Association for Computational Linguistics",
    url = "https://aclanthology.org/2024.acl-long.739/",
    doi = "10.18653/v1/2024.acl-long.739",
    pages = "13679--13707"
}

@article{scott1962decision,
  title   = {A decision method for validity of sentences in two variables},
  author  = {Scott, Dana},
  journal = {Journal of Symbolic Logic},
  volume  = {27},
  pages   = {477},
  year    = {1962}
}

@article{mortimer1975,
	author = {Michael Mortimer},
	doi = {10.1002/malq.19750210118},
	journal = {Mathematical Logic Quarterly},
	number = {1},
	pages = {135--140},
	publisher = {Wiley-Blackwell},
	title = {On Languages with Two Variables},
	volume = {21},
	year = {1975}
}

@book{kalishMontague1964,
  title     = {{Logic: Techniques of Formal Reasoning}},
  author    = {Kalish, Donald and Montague, Richard},
  year      = {1964},
  publisher = {Harcourt, Brace \& World},
  address   = {New York}
}

@InProceedings{10.1007/978-3-540-78800-3_24,
author="de Moura, Leonardo
and Bj{\o}rner, Nikolaj",
editor="Ramakrishnan, C. R.
and Rehof, Jakob",
title="Z3: An Efficient SMT Solver",
booktitle="Tools and Algorithms for the Construction and Analysis of Systems",
year="2008",
publisher="Springer Berlin Heidelberg",
address="Berlin, Heidelberg",
pages="337--340",
isbn="978-3-540-78800-3"
}

@article{DavisPutnam1960,
author = {Davis, Martin and Putnam, Hilary},
title = {A Computing Procedure for Quantification Theory},
year = {1960},
publisher = {Association for Computing Machinery},
address = {New York, NY, USA},
volume = {7},
number = {3},
doi = {10.1145/321033.321034},
journal = {J. ACM},
month = {July},
pages = {201–215}
}

@inproceedings{huang-chang-2023-towards,
    title = "Towards Reasoning in Large Language Models: A Survey",
    author = "Huang, Jie  and
      Chang, Kevin Chen-Chuan",
    editor = "Rogers, Anna  and
      Boyd-Graber, Jordan  and
      Okazaki, Naoaki",
    booktitle = "Findings of the Association for Computational Linguistics: ACL 2023",
    month = jul,
    year = "2023",
    address = "Toronto, Canada",
    publisher = "Association for Computational Linguistics",
    url = "https://aclanthology.org/2023.findings-acl.67/",
    doi = "10.18653/v1/2023.findings-acl.67",
    pages = "1049--1065"
}

@inproceedings{
mondorf2024beyond,
title={Beyond Accuracy: Evaluating the Reasoning Behavior of Large Language Models - A Survey},
author={Philipp Mondorf and Barbara Plank},
booktitle={First Conference on Language Modeling},
year={2024},
url={https://openreview.net/forum?id=Lmjgl2n11u}
}

@article{Chen2026,
  author  = {Chen, Qiguang and Qin, Libo and Liu, Jinhao and Peng, Dengyun and
             Guan, Jiannan and Wang, Peng and Hu, Mengkang and Zhou, Yuhang and
             Gao, Te and Che, Wanxiang},
  title   = {Towards reasoning era: a survey of long chain-of-thought for reasoning large language models},
  journal = {Science China Information Sciences},
  year    = {2026},
  volume  = {69},
  number  = {6},
  pages   = {161101},
  doi     = {10.1007/s11432-025-4665-8},
  url     = {https://doi.org/10.1007/s11432-025-4665-8},
  issn    = {1869-1919}
}

@article{yang2023logical,
  title={Logical reasoning over natural language as knowledge representation: A survey},
  author={Yang, Zonglin and Du, Xinya and Mao, Rui and Ni, Jinjie and Cambria, Erik},
  journal={arXiv preprint arXiv:2303.12023},
  year={2023}
}

\appendix
\section{Lexicon and Sentence Types}
\label{app:sentences}

\textsc{LogicSkills} sentences span a range of first-order constructions. Table~\ref{tab:sentence-types} gives representative examples, covering atomic predicates, quantification (including nested and anaphoric forms), Boolean connectives, and conditionals. Each schema admits systematic surface variants (e.g., VP ellipsis, contrastive conjunctions, unless-disjunctions) that preserve the same logical structure. 

Each sentence is realized in two surface languages: controlled English and a Carroll-style nonsense language, each with its own fixed lexicon. The fixed lexicons contain 3 constants (names) and 13 predicates, which bounds the maximum vocabulary size of any sentence.

\section{Dataset and Task Construction Pipeline}
\label{app:pipeline}

\paragraph{Sentence generation.}
All tasks are grounded in a large shared sentence inventory (100k+ sentences) generated compositionally. Each sentence is paired with a surface realization (English or Carrollian), an FO$^{2}$ logical form, its abstract syntax tree (AST), and a schema of abbreviation (SOA) mapping logical symbols to lexical items. English and Carroll realizations are linked by a counterpart relation preserving logical form. Sentences are labeled by syntactic type (atomic, negated, conditional, quantified, etc.) and stored in a database with fields for surface form, SOA, FO$^{2}$ formula, AST, language tag, and a counterpart identifier linking English and Carroll realizations.

\paragraph{Argument generation.}
Arguments are generated by sampling small premise sets (3--5 sentences) from the sentence table and applying solver-based filtering. Premises must be jointly satisfiable, and candidate conclusions are tested for entailment using Z3. To enforce non-triviality, we exclude theorems, conclusions derivable from domain constraints alone, and conclusions derivable from only a proper subset (at most 20\%) of the premises. For each valid argument, five invalid arguments are constructed using alternative conclusions that fail to follow; distractors are constrained to share subject matter with the valid conclusion (same predicates and constants), preferentially match its syntactic type, and be solver-verified as non-entailed. Arguments are stored in a database with fields for premise identifiers, conclusion identifier, validity label, and language, and form the large shared pool from which validity and countermodel tasks are derived.

\paragraph{Task instantiation.}
Three task families are derived from the sentence and argument tables. Formal symbolization tasks are constructed from individual sentences paired with their SOA, with the associated FO$^{2}$ formula as the target. Although the fixed lexicons permit a larger vocabulary in principle, in practice sentences used to instantiate tasks involve only 3--4 predicates and 1--2 constants (4--6 symbols total). Argument validity tasks are constructed from solver-certified valid arguments, each paired with five invalid alternatives sharing the same premises. Countermodel construction tasks are constructed from solver-certified invalid arguments and require a finite model that satisfies all premises while falsifying the conclusion.

\paragraph{Countermodel domains.}
The countermodel task fixes the domain to $\{0,1,2\}$. Solver-certified non-entailment guarantees only that some countermodel exists, of unbounded size, so we verified that each item is solvable within this
domain: we ran Schwarz's tree proof generator (tpg; \citealp{schwarz2025tpg}) on every invalid argument admitted to the task and recorded the size of the countermodel returned. The maximum domain size observed across all 300 items was 3, confirming that every item is solvable within the fixed domain.

\section{Evaluation Procedure}
\label{app:eval}
All task outputs are verified automatically using Z3. The three tasks differ in how correctness is defined. In the validity task, a model is given a premise set and six candidate conclusions and must return the indices of those conclusions that follow from the premises. The gold answer is the set of such indices, which may be empty, a singleton, or multi-element; scoring is set equality, so omitting an entailed conclusion and including a non-entailed one are both errors. The symbolization and countermodel tasks admit multiple correct answers, so correctness is defined semantically rather than syntactically.

In the formal symbolization task, accuracy is defined as producing a formula logically equivalent to the target representation. Two formulas are considered equivalent if they have the same truth conditions in all structures. Evaluation therefore relies on logical equivalence checking rather than string identity. Natural-language sentences may admit multiple logically equivalent formalizations. For example, the formulas $\forall x(Dx \rightarrow \neg Bx)$ and $\neg \exists x(Dx \wedge Bx)$ are logically equivalent and therefore treated as equally correct representations of the same sentence.

Invalid arguments may admit many countermodels. The countermodel task therefore defines correctness existentially: a candidate structure is correct if it makes all premises true and the conclusion false. The task fixes a small finite domain and requires a fully specified structure over that domain. Candidate structures are normalized into domain assignments, constant interpretations, and predicate extensions prior to verification. Any structure satisfying these semantic conditions is accepted as correct.

Code for dataset generation, task construction, and evaluation is available at \url{https://github.com/brianrabern/LogicSkills}.

\section{Error Analysis Methodology}\label{app:err_analysis}
Our evaluation pipeline produced structured outputs beyond binary correctness labels, enabling systematic error analysis across all tasks. For each instance, we retained the raw model response, the extracted or parsed answer, the gold-standard answer, and task-specific diagnostic metadata from the evaluator.

\paragraph{Formal symbolization} For \emph{symbolization}, each response included an assessment label (e.g., \emph{Not logically equivalent}, \emph{Failed to parse}). Incorrect outputs were filtered and categorized by matching these labels, allowing semantic errors to be distinguished from syntactic or structural failures.

\paragraph{Countermodel construction} For \emph{countermodel construction}, the model checker returned a list of diagnostic error messages (e.g., \emph{Not a countermodel}, \emph{Missing interpretation for:\! X}, \emph{Binary predicate `P' must be a list of pairs}, etc.). Failures were categorized by matching error-message patterns, with additional checks performed by evaluating premises and conclusions against the proposed structure when needed.

\paragraph{Argument validity} For \emph{validity}, both the extracted and correct answers were represented as sets of statement indices. Set comparisons were used to distinguish incorrect selections from \emph{superset} errors in which the correct answer was included alongside additional invalid options.

All error analyses were performed programmatically over the full evaluation set, and the resulting error-type distributions correspond to those reported in the main text.

\section{Meta-Evaluation of Answer Extraction}
\label{app:meta-eval}

Raw model responses were post-processed by an extractor LLM to normalize outputs into task-specific formats prior to scoring. To assess the reliability of this extraction step independently of task correctness, we conducted a meta-evaluation.

We randomly sampled 100 evaluation instances, balanced across all task types (symbolization, validity, and countermodel construction) and languages where applicable, and including both correct and incorrect model responses. For each instance, the evaluator was given the original task prompt, the model’s raw response, and the extracted answer produced by the extractor.

The meta-evaluator (GPT-4o) rendered a binary judgment—faithful or unfaithful—indicating whether the extracted answer accurately reflected the content of the raw response, independent of correctness. Five cases were flagged as potentially unfaithful.

All flagged cases, together with a random sample of 10 unflagged instances, were subsequently reviewed by a human evaluator (a meta-meta-evaluation). Of the five flagged cases, one constituted a clear extraction error, in which the model revised a predicate assignment mid-response and the extractor failed to capture the update. One case involved a normalization failure, where a malformed but correctly extracted formula was not repaired. The remaining three cases involved internally inconsistent raw responses (e.g., conflicting predicate arities or self-contradictory conclusions); in these cases, the extractor’s output was judged reasonable given the ambiguity of the source response.

After human adjudication, only one instance was deemed a clear extraction failure, with one additional case judged ambiguous, corresponding to an effective extraction accuracy of 98--99\%.

\section{Task Prompts}\label{sec:appendix_prompts}
Figures~\ref{fig:argument-validity} to~\ref{fig:countermodel} illustrate the prompts used for each task type described in Section~\ref{subsec:task_types}.

\section{Additional Results}\label{sec:appendix_results}
Table~\ref{tab:full_overview} reports the complete per-model results underlying \S\ref{sec:results}.

\section{Use of AI Assistants}\label{sec:appendix_ai_tools}
We used GitHub Copilot to assist with parts of the project's source code and Claude for proofreading and copy-editing of the manuscript. All research design, analysis, and claims are the authors' own, and all references were verified manually.

\newpage
\begin{table*}[b!]
\centering
\begin{tabular}{lccccc}
\hline
\textbf{Model} & \textbf{Val (Car.)} & \textbf{Val (Eng.)} & \textbf{Sym (Car.)} & \textbf{Sym (Eng.)} & \textbf{Counter} \\
\hline
meta\_llama-3.1-8B-instruct   & 0.46 & 0.45 & 0.14 & 0.14 & 0.00 \\
meta\_llama-3.1-70B-instruct  & 0.84 & 0.86 & 0.66 & 0.69 & 0.12 \\
qwen2.5-math-72B-instruct     & 0.80 & 0.80 & 0.80 & 0.80 & 0.15 \\
anthropic\_claude-3.7-sonnet  & 0.96 & 0.94 & 0.71 & 0.76 & 0.42 \\
openai/gpt-4o                 & 0.87 & 0.85 & 0.74 & 0.72 & 0.10 \\
google\_gemini-2.5-flash      & 0.96 & 0.94 & 0.76 & 0.79 & 0.13 \\
deepseek\_chat                & 0.81 & 0.76 & 0.80 & 0.81 & 0.35 \\
microsoft\_phi-4              & 0.80 & 0.73 & 0.51 & 0.57 & 0.14 \\
\hline
\textbf{qwen3-32B}            & \textbf{0.95} & \textbf{0.97} & \textbf{0.82} & \textbf{0.85} & \textbf{0.89} \\
\textbf{openai/o3}            & \textbf{0.99} & \textbf{0.99} & \textbf{0.97} & \textbf{0.94} & \textbf{0.99} \\
\hline
\end{tabular}
\caption{Accuracy across tasks and models (300 problems per setting). 95\% Wilson intervals are at most $\pm5.6$ points wide.}
\label{tab:full_overview}
\end{table*}

\newpage
\begin{figure*}[b!]
\begin{mdframed}[backgroundcolor=gray!5, linewidth=0pt, innerleftmargin=10pt, innerrightmargin=10pt]
\textbf{\textsc{Argument Validity Prompt}}

\vspace{0.6em}
{\ttfamily

\noindent Your task is to solve a logical reasoning problem. Use any approach you find effective, but clearly and explicitly state your final answer.

\vspace{0.6em}

\textbf{Task}

\vspace{0.2em}

Consider the following situation:
\begin{quote}
Everything is a tove, a borogove, or a rath (exclusively), and there's at least one of each. Zindle or Bungo will whiffle. Only toves will whiffle. Every rath chortled at Bungo. If Zindle will whiffle, then every rath is mimsy only if every borogove chortled at Bungo.
\end{quote}

\noindent Which, if any, of the following statements must be true in this situation?

\vspace{0.4em}
\begin{enumerate}
  \setlength{\itemsep}{2pt}
  \setlength{\topsep}{2pt}
  \item Zindle and Bungo will whiffle.
  \item Every tove will whiffle.
  \item Not all toves will whiffle.
  \item A tove will whiffle.
  \item No toves will whiffle.
  \item Zindle will whiffle.
\end{enumerate}
}
\end{mdframed}
\caption{Illustrative instance of the \emph{Argument Validity} task.}
\label{fig:argument-validity}
\end{figure*}

\begin{figure*}[tbp]
% \small 
\begin{mdframed}[backgroundcolor=gray!3, linewidth=0pt, innerleftmargin=10pt, innerrightmargin=10pt]
\textbf{\textsc{Formal Symbolization Prompt}}

\vspace{0.6em}
 \ttfamily
\noindent Your task is to translate the provided sentence into formal predicate logic, using the abbreviations provided.

\vspace{0.6em}
\textbf{Instructions}
\vspace{-0.6em}
\begin{itemize} \setlength{\parskip}{0pt} 
  \item Use only the abbreviations given.
  \item Return a single well-formed formula in standard predicate logic syntax.
  \item Use standard logical symbols:
  \begin{itemize}\setlength{\parskip}{0pt} 
    \item Quantifiers: $\forall$, $\exists$
    \item Connectives: $\neg$, $\wedge$, $\vee$, $\rightarrow$, $\leftrightarrow$
    \item Do \emph{not} include any explanation or extra text---just the formula.
  \end{itemize}
\end{itemize}

\vspace{0.6em}
\textbf{Example}

\vspace{0.2em}
Sentence: Every linguist admires Charlie.\\
Abbreviations: 
\vspace{-0.6em}
\begin{itemize} \setlength{\parskip}{0pt} 
\item L: “[1] is a linguist”
\item R: “[1] admires [2]”
\item c: “Charlie”\\
\end{itemize}
\vspace{-0.6em}
Translation: $\forall x(Lx \rightarrow Rxc)$

\vspace{0.5em}
\hrule
\vspace{1em}

\textbf{Task}
\vspace{0.2em}

\noindent Sentence: Not all raths will whiffle and no toves chortled at Bungo.

\vspace{0.5em}
\noindent Abbreviations:
\vspace{-0.6em}
\begin{itemize} \setlength{\parskip}{0pt} 
  \item O: “[1] is a rath”
  \item F: “[1] will whiffle”
  \item M: “[1] is a tove”
  \item P: “[1] chortled at [2]”
  \item c: “Bungo”
\end{itemize}
\end{mdframed}
\caption{Illustrative instance of the \emph{Formal Symbolization} task. }
\label{fig:formal-symbolization}
\end{figure*}

\begin{figure*}[tbp]
% \small 
\begin{mdframed}[backgroundcolor=gray!5, linewidth=0pt, innerleftmargin=10pt, innerrightmargin=10pt]
\textbf{\textsc{Countermodel Construction Prompt}}

\vspace{0.6em}
{\ttfamily
\noindent Show that the provided argument is invalid by giving a countermodel---one where all premises are true and the conclusion is false.

\vspace{0.6em}
\textbf{Instructions}
\vspace{-0.4em}
\begin{enumerate}
  \setlength{\itemsep}{2pt}
  \setlength{\topsep}{2pt}
  \item Provide assignments for all constants and predicates used in the argument.
  \item Respect predicate arity:
  \begin{itemize}
    \setlength{\itemsep}{1pt}
    \item Monadic predicates take one argument (e.g., $Mx$).
    \item Binary predicates take two arguments (e.g., $Pxy$).
  \end{itemize}
  \item Use the fixed domain $[0, 1, 2]$.
\end{enumerate}

\vspace{0.6em}
\textbf{Required Format}
\vspace{-0.4em}
\begin{itemize}
  \setlength{\itemsep}{2pt}
  \item Domain: $[0, 1, 2]$
  \item Constants: ``a": $0$
  \item Monadic predicates: ``F": $[0, 2]$
  \item Binary predicates: ``R": $[[0, 1], [2, 0]]$
\end{itemize}

\vspace{0.5em}
\hrule
\vspace{1em}

\textbf{Argument}
\vspace{0.8em}

\begin{flalign*}
& \big(\forall x \,(Mx \vee (Nx \vee Ox)) \;\wedge\;
   \big(\neg \exists x \,(Mx \wedge Nx) \;\wedge\;
     (\neg \exists x \,(Mx \wedge Ox) \;\wedge\; 
      \neg \exists x \,(Nx \wedge Ox))\big)\big), \\[6pt]
& (\exists x \,Mx \;\wedge\; (\exists x \,Nx \;\wedge\; \exists x \,Ox)), \\[6pt]
& \neg \exists x \exists y \,((Mx \wedge My) \wedge Pxy), \\[6pt]
& (\neg \exists x \,(Ox \wedge Qxc) \rightarrow Mb), \\[6pt]
& (\neg \exists x \,(Nx \wedge Rxc) \rightarrow \exists x \,(Nx \wedge \neg Rxb)), \\[6pt]
& (\forall x \,(Ox \rightarrow Rxc) \rightarrow (Mb \rightarrow \forall x \,(Nx \rightarrow Pxc))), \\[6pt]
& (Pcc \wedge \forall x \,(Ox \rightarrow Rxa)) \\[10pt]
& \vDash Mc
\end{flalign*}
}
\end{mdframed}
\caption{Illustrative instance of the \emph{Countermodel Construction} task.}
\label{fig:countermodel}
\end{figure*}

\newpage

\begin{table*}[b!]
\setlength{\tabcolsep}{5pt}
\centering
\small
\begin{tabular}{l p{4cm} p{4cm}}
\hline
\textbf{Sentence type} & \textbf{English example} & \textbf{Carroll example} \\
\hline
atomic\_monadic
& Hazel drank.
& Bungo gyred. \\

atomic\_dyadic
& Hazel kicked Lewis.
& Bungo galumphed over Rafin. \\

negation
& Lewis isn’t happy.
& Rafin isn’t uffish. \\

quantified\_universal\_affirmative
& Every monkey is asleep.
& Every rath is mimsy. \\

quantified\_particular\_affirmative
& A donkey will run.
& A tove will whiffle. \\

quantified\_universal\_negative
& No humans chased Alfred.
& No borogoves snicker-snacked Zindle. \\

quantified\_particular\_negative
& Not all donkeys chased Alfred.
& Not all toves snicker-snacked Zindle. \\

quantified\_only\_restrictor
& Only humans will run.
& Only borogoves will whiffle. \\

quantified\_name\_all
& Hazel chased every human.
& Bungo snicker-snacked every borogove. \\

quantified\_name\_some
& Alfred chased a donkey.
& Zindle snicker-snacked a tove. \\

quantified\_all\_all
& Every monkey kicked every donkey.
& Every rath galumphed over every tove. \\

quantified\_all\_all\_all
& Every donkey saw every donkey that saw every monkey.
& Every tove chortled at every tove that chortled at every rath. \\

quantified\_all\_all\_back
& Every donkey chased every monkey that chased it.
& Every tove snicker-snacked every rath that snicker-snacked it. \\

quantified\_all\_some
& Every monkey kicked a human.
& Every rath galumphed over a borogove. \\

quantified\_all\_some\_back
& Every human saw a donkey that saw it.
& Every borogove chortled at a tove that chortled at it. \\

quantified\_some\_all
& A monkey chased every human.
& A rath snicker-snacked every borogove. \\

quantified\_some\_all\_back
& A monkey saw every human that saw it.
& A rath chortled at every borogove that chortled at it. \\

quantified\_some\_some
& A human chased a donkey.
& A borogove snicker-snacked a tove. \\

quantified\_some\_some\_back
& A donkey kicked a monkey that kicked it.
& A tove galumphed over a rath that galumphed over it. \\

quantified\_some\_some\_some
& A human kicked a human that kicked a monkey.
& A borogove galumphed over a borogove that galumphed over a rath. \\

quantified\_no\_all
& No humans kicked every human.
& No borogoves galumphed over every borogove. \\

quantified\_no\_some
& No humans chased a human.
& No borogoves snicker-snacked a borogove. \\

quantified\_no\_some\_back
& No donkeys kicked a monkey that kicked it.
& No toves galumphed over a rath that galumphed over it. \\

quantified\_rev\_some\_all
& There is a donkey that every monkey kicked.
& There is a tove that every rath galumphed over. \\

quantified\_rev\_no\_all
& There is not a monkey that every human chased.
& There is not a rath that every borogove snicker-snacked. \\

quantified\_some\_self
& A donkey kicked itself.
& A tove galumphed over itself. \\

conjunction\_simple
& Hazel saw Lewis and not all monkeys kicked Lewis.
& Bungo chortled at Rafin and not all raths galumphed over Rafin. \\

disjunction\_simple
& Alfred drank or a monkey kicked Alfred.
& Zindle gyred or a rath galumphed over Zindle. \\

conditional\_if\_then
& If Hazel is a human, then Alfred isn’t a human.
& If Bungo is a borogove, then Zindle isn’t a borogove. \\

biconditional\_just\_in\_case
& Hazel is happy just in case every donkey chased Lewis.
& Bungo is uffish just in case every tove snicker-snacked Rafin. \\
\hline
\end{tabular}
\caption{
Sentence types used in \textsc{LogicSkills} with representative English and Carroll examples.}
\label{tab:sentence-types}
\end{table*}

\end{document}